\documentclass[letterpaper]{article}
\usepackage{natbib, alifeconf}
\usepackage{array}
\usepackage{hyperref}

\title{Host-Pathongen Co-evolution Inspired Algorithm Enables Robust GAN Training}
\author{Andrei Kucharavy$^{1}$, El Mahdi El Mhamdi$^{1}$ \and Rachid Guerraoui$^{1}$ \\
\mbox{}\\
$^1$Ecole Polytechnique Federale of Lausanne, Route Cantonale 1015, Lausanne \\
andrei.kucharavy@gmail.com}

\begin{document}
\maketitle

\begin{abstract}
Generative adversarial networks (GANs) are pairs of artificial neural networks that are trained one against each other. The outputs from a generator are mixed with the real-world inputs to the discriminator and both networks are trained until an equilibrium is reached, where the discriminator cannot distinguish generated inputs from real ones.
Since their introduction, GANs have allowed for the generation of impressive imitations of real-life films, images and texts, whose fakeness is barely noticeable to humans. 
Despite their impressive performance, training GANs remains to this day more of an art than a reliable procedure, in a large part due to training process stability. Generators are susceptible to mode dropping and convergence to random patterns, which have to be mitigated by computationally expensive multiple restarts.
Curiously, GANs bear an uncanny similarity to a co-evolution of a pathogen and its host's immune system in biology. In a biological context, the majority of  potential pathogens indeed never make it and are kept at bay by the hots' immune system. 
Yet some are efficient enough to present a risk of a serious condition and recurrent infections.
Here, we explore that similarity to propose a more robust algorithm for GANs training.
We empirically show the increased stability and a better ability to generate high-quality images while using less computational power.
\end{abstract}

\section{Introduction}

Since the introduction of Generative Adversarial Networks (GANs) in \cite{goodfellow2014generative}, they have been touted as means to produce life-like images, films, sounds and texts (\cite{goodfellow2016nips}). Latest versions of GANs have reached a level where it becomes virtually impossible for the general public to distinguish their products from their real-world counterparts: eg. portrait photos in \cite{karras2019analyzing}, artistic paintings \cite{gatys2015neural} or text generation from a prompt \cite{radford2019language}. Besides their applications in image and text, GANs have been seen as promising means to better understand complex systems, such as for instance  molecular mechanisms linking gene expression in cells to their phenotype, as for instance in \cite{eraslan2019deep}.

Despite these impressive achievements, GANs still present critical problems upon their training. Two notable problems are mode collapse and vanishing gradients \cite{arjovsky2017wasserstein, arora2017generalization, salimans2016improved}. The first problem - mode collapse - consists in the generator becoming biased to producing a single class of images or a single image instance as a result of that specific class being more confusing to the discriminator. The second problem - vanishing gradients - is due to the generator updates becoming less and less significant due to the discriminator becoming good at discriminating between real images and the ones that were generated artificially.

Tweaks to GANs training modes that would address this issue have been suggested. Some of the most notable are Wasserstein GANs introduced in \cite{arjovsky2017wasserstein}, Gradient Norm regularized GAN from \cite{salimans2016improved}, Least Square GAN suggested in \cite{mao2017least} and autoencoder-aided GANs introduced in \cite{berthelot2017began}. Unfortunately, a more in-depth exploration of their performance in \cite{lucic2018gans} have shown that their differences are minor and mostly due to the number of the GAN training restarts and selection of best-performing instances.

The similarity of this observation with the results in pathogen-host co-evolution are striking. In their work, \cite{hermisson2005soft} have shown that pathogens adaptation to hostile environment - such as drugs or primed host immune system are not due as much to the evolution after the hostile environment application, but rather to a pre-existing diversity creating a large pool that would contain potentially adaptive variation and upon which evolutionary processes could build upon. A diversity of pre-existing pool is so important that, along with other pathogens, cancer cells use a separate mechanism to generate diversity and rapidly adapt to chemotherapies, senescence and acquire spreading abilities \cite{kucharavy2018robustness}.

If we build on the pathogen-host analogy, we can as well notice that some pathogens can evolve much more rapidly and present a significantly higher danger to their hosts. Specifically, single-strand RNA viruses (ssRNA viruses) evolve rapidly, to the point that some of them (notably influenza A) can recurrently escape detection by host immune system and re-infect hosts that have previously developed immunity to them (\cite{klingen2018silico, nelson2007evolution}). For some of them, the mutation rate is so high, that the concept of a species becomes porous and can be abandoned in favor of treating each separate viral infection as a separate quasi-species (\cite{domingo2019viral}). Viruses that can cross between species (zoonotic viruses) are frequent and can be particularly dangerous, given that new host populations have no immunity or cross-immunity to it, and their immune system would not be able to keep pace with the infection (\cite{holmes2009evolutionary, mollentze2020viral}). A significant proportion of most dangerous viruses in humans are specifically ssRNA zoonotic viruses. Ebola, Marburg, Hantavirus, Lassa Fever, Rabies, Middle-Eastern Respiratory Syndrome (MERS), Measles, H5N1 avian flu or H1N1 Spanish Influenza are all caused by ssRNA zoonotic viruses and, combined with yet unknown ssRNA zoonotic viruses, pose a serious threat to public health (\cite{carrasco2017rna}).

Recent works in the evolutionary theory have shown that simplistic models of natural evolution introduced during the modern evolutionary synthesis by \cite{fischer1930genetical} and later refined by \cite{kimura1983neutral, ohta1992nearly}, are equivalent to a an ergodic walk in genotypic space. Rare beneficial mutations and more frequent mostly neutral ones allow transition from one highly optimized state to another, even more highly optimized state, as discussed in \cite{gillespie1994causes}. Most notably, many critical properties of this walk do not depend on specific mechanisms by which transitions to a more optimized state occur - just on their high-level statistical properties, that are frequently satisfied in optimization tasks, as described in \cite{orr2005theories, orr2005genetic}. There is however an important caveat. Such adaptations require a changing fitness landscape to lead to a persistent, long-term adaptations accumulation. Without it, evolutionary process gets stuck in local fitness peaks \cite{kauffman1987towards, gillespie1984molecular, orr2002population}. As \cite{gould1972punctuated} noted, without environment change, biological systems, once adapted to it, would remain unchanged for hundreds of millions of years - akin to horseshoe crabs.

Such a generality implies that we would expect the results from the theory of evolution to be applicable to the co-evolutionary pairs of generators and discriminators.
Unfortunately, specifying the evolutionary process that performs well is a non-trivial task. Here, instead we will focus on one specific aspect - the effect of the transition between heterogeneous naive populations.

We are far from the first ones to apply the evolutionary theory to GANs. So far the efforts have focused in two different directions. 
First - using random perturbations and recombination events in existing generator-discriminator pairs positioned in a spatial grid in order to maintain diversity while optimizing fitness, most recently represented by \cite{wang2019evolutionary, toutouh2019spatial}. Second - search for optimal generator/discriminator artificial neural networks architectures, spearheaded notably by \cite{ren2019eigen}.

The novelty of our approach consists in assuming that, due to the speed of RNA viruses adaptive evolution and the speed of the immune system antigen generation, we can assume that a single training epoch is equivalent to a bout of adaptive evolution of virus and an attempt of immune system to learn to suppress the virus as it is evolving. We adopt a host-pathogen view of co-evolution - a field that has been recently developed from a theoretical point of view to better understand antibiotics resistance emergence in bacteria (\cite{wilson2016population}), tri-therapy resistance emergence in HIV (\cite{hermisson2005soft}) and rapid acquisition of invasive phenotype and drug resistance in cancers (\cite{kucharavy2018robustness}).

\section{Methods}

\subsection{Model scope}

Within the scope of our model, we assume to not be dealing with major changes to the pathogen envelope structure or inner workings, but rather clades.\footnote{Sets of distinct, closely related sub-groups in a single species.} undergoing adaptive antigen evolution. In other terms - a group of quasi-species, with strong constraints on genome architecture, but with a variation in epitope-coding genes. This means that in GANs, we do not search for optimal structures, but instead focus on a predefined set of generator/discriminator neural network architectures.

Our hypothesis is that by providing several naive populations of discriminators and using a population of generators that would train on all of them, we will be able to preserve enough pathogen heterogeneity and by several jumps through naive heterogeneous populations we will be able to force several adaptive swipes and avoid getting trapped in local fitness optima. For such training structures, we expect to achieve better results than by performing long bouts of training of single generator-single discriminator pairs with multiple restarts.

\subsection{Computation framework}

In order to perform computational experiments, we use of Pytorch v 1.4 (\cite{paszke2019pytorch}) and Python 3.8.1. The full code needed to replicate the results in this paper is available at \url{https://github.com/chiffa/GAN_evo/releases/tag/v0.1.1}, commit 0ed835a.

In our experiments, we focus on the GANs trained on the MNIST dataset (\cite{lecun1998gradient}).

\subsection{Discriminators and generators architecture}

For the generator, we adopt a commonly used five transpose-convolutional layers architecture with batch normalization between the layers. For discriminators, we also adapt a commonly used five convolutional layers with batch normalization architecture.

\begin{figure}[!htb]
\begin{center}
\includegraphics[width=3.0in]{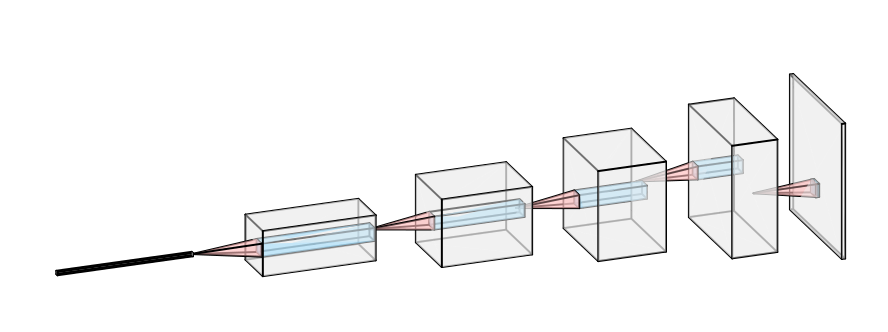}
\caption{Generator architecture. Each convolution transpose layer is of kernel=4, stride=2 and padding=1, followed by batch normalization and a ReLU non-linear activation layer.}
\label{gen_architecture}
\end{center}
\end{figure}

\begin{figure}[!htb]
\begin{center}
\includegraphics[width=3.0in]{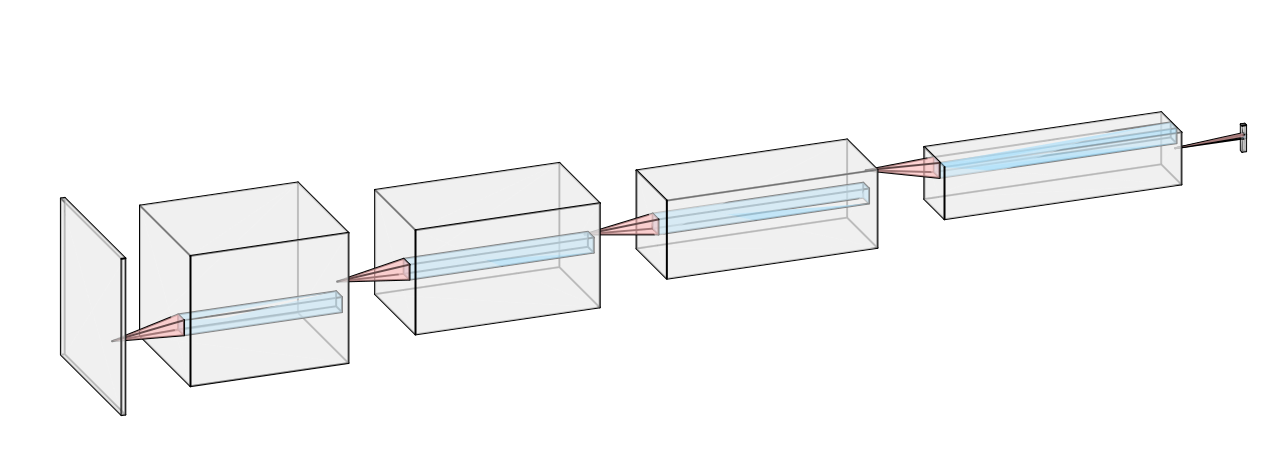}
\caption{Discriminator architecture. Each convolution layer is of kernel=4, stride=2 and padding=1, followed by batch normalization and a LeakyReLU/PreLU non-linear activation layer.}
\label{disc_architecture}
\end{center}
\end{figure}

We refer to the discriminator architecture above as "base". In order to simulate the immune system heterogeneity between different species, we adapt two variants of it: "light", where the number of features was divided by 2 (down to 32 from 64), and "PreLU" where non-linear layers of ReLU have been changed to PreLU layers. More optimal architectures are possible and can be found using evolutionary methods, as has been described previously in \cite{ren2019eigen}, but are outside the scope of this paper.

\subsection{Frechnet inception score}

Frechnet Inception Score (FID) between two images (real or generated) is defined as the Frechnet distance between the activation vector of the Inception v3 computer vision models and was introduced in \cite{heusel2017gans}. As such, it has been shown to correlate well with the similarity between images as perceived by human users. It has also been suggested to be an optimal estimator for the perceived quality of GAN outputs and a golden standard to which the GAN training performance is to be estimated, after accounting for computational expense including training restarts.

\subsection{Baseline comparison}

Following the \cite{heusel2017gans}, we consider the standard approach to training GANs is to train them in separate, non-interacting pairs, then choose the one with the highest FID score.

Due to that, we compare training methods by computing statistics on the best FID achieved per run. Each run has a pre-defined number of restarts and is limited by a computational budget - the number of epochs available for training.

\subsection{Host-Pathogen Fitness}

In the implementations where the generators and discriminators undergo evolutionary selection, we define their fitness scores consistently with prior research in \cite{kucharavy2018robustness}, as a Weibull distribution. The shape parameter $k$ of the Weibull distribution corresponds to the the number of effective phenotype dimensions in which the evolution takes place. The penalty to the hosts for failing to properly classify real images as such ($err_{real}$) is controlled by the "autoimmunity" factor $a$. The failure to recognize the images generated by a generator $err_{gen}$  as such is penalized with the "viral reproduction" factor $v$, indicating the amount of organism's resources exploited by the pathogen to reproduce and the maximum number of pathogen replicas created\footnote{Please note that the viral reproduction factor here is different from the $R_0$ in epidemiology.}. 

If the pathogen fitness (how well it is able to reproduce itself) within host is more than $1.$, we consider that the host is infected with pathogen and the host fitness is now affected by the viral load of all the pathogens it is infected with. We assume that the the autoimmunity and pathogen loads are all independent. We also assume that the host immune system will not efficiently learn to recognize the pathogen if the fitness deficit imparted by the pathogen is below 5\%. Finally, we assume that the host fitness has a lower bound of 1\% of the maximal possible fitness. 

We assume that all the negative effects of the pathogen are due to its use of host's resources to reproduce. Due to that, the fitness of the pathogen is proportional to the ability of its generator to mislead the host's discriminator, controlled by the same "virality" factor $v$. Just as for the discriminator, the fitness for the pathogen is assumed to have a lower bound of 0.25\% of the maximal possible fitness.

In this context, we can write fitness functions as:
$$
F_{host} = 1 - Weibull_{CDF}(k, \sqrt{a^2 \cdot err_{real}^2+v^2 \cdot err_{gen}^2})
$$

$$
F_{path} = v \cdot Weibull_{CDF}(k, v \cdot err_{gen})
$$

\subsection{Pathogen Propagation Structures}
In our tests of different GAN training architectures, we used the following configurations:

\textbf{Standard Round-Robin:} In this architecture, we use a uniform population of hosts and pathogens and perform a complete cross-match between all the generators and discriminators, until we've exhausted the epochs budget. This architecture neither uses fitness nor performs selection. From the pathogen propagation perspective, it is equivalent to a propagation of pathogens in a static ring, one after another, trying to adapt to the host's cross-immunity with other pathogens. The specific implementation here uses three rounds of 5 pathogens and 5 hosts, for a total budget of 75 epochs per run.

\textbf{Stochastic Round-Robin:} In this architecture, we use a uniform population of hosts and pathogens and randomly select the generators and discriminators to match in training, until the computational budget is exhausted. This architecture also neither uses fitness nor performs selection. From the pathogen propagation point of view, it is equivalent to a circulation of pathogens at random in a population. The specific implementation here performs a first round of deterministic Round-Robin and then a total of 50 epochs of randomized transmission for a total budget of 75 epochs per run.

\textbf{Round-Robin with population jumps:} In this architecture, we use a set of uniform populations of hosts. Pathogens and hosts from a single population perform a complete cross-match (static ring round), after which they all jump to a new population, naive to the pathogens and re-perform a complete cross-match. This architecture also neither uses fitness nor performs selection. The specific implementation here uses three naive populations of 5 hosts and 5 pathogens and two additional rounds for the last population, for a total budget of 125 epochs per run.

\textbf{Round-Robin with heterogeneous population jumps:} This architecture is similar to the one above, except the populations of hosts correspond to different discriminator architectures, in order: 'light', 'PreLU', 'base'. The specific implementation here uses 5 hosts from each architecture and 5 pathogens and two additional rounds for the last population, for a total budget of 125 epochs per run.

\textbf{Evolution with heterogeneous populations jumps:} This architecture is similar to the one above, except that after a deterministic cross-match took place in a population, a stochastic, fitness-based chain of transmissions takes place in the same population. A pathogen and a host are selected at random, but with probabilities proportional to their fitness and are matched one against each other. The evolutionary process continues as long as the epoch budget is available. The specific implementation here uses 3 hosts from each architecture and 4 pathogens and a budget of 12 epochs per evolutionary step, for a total budget of 72 epochs per run.

\subsection{Normalization and baseline comparison}
We consider that the baseline approach for finding the best GAN for the available computational budget is to train a number of GANs for a set number of epochs and select the generator with the lowest FID. Here, we chose as the baseline training GANs for 15 epochs and a population size 10 - a budget of 150 epochs.

In order to collect results, we perform at least 5 runs of each method and compute the median, mean and standard deviation of the best FIDs achieved per run in each method. We then compare the medians of each method, in order to decrease the impact of outliers on the representative FID, and compute the Student t-statistic to determine whether the difference between methods is statistically significant.

\section{Results}

\begin{figure}[!htb]
\begin{center}
\includegraphics[width=3.0in]{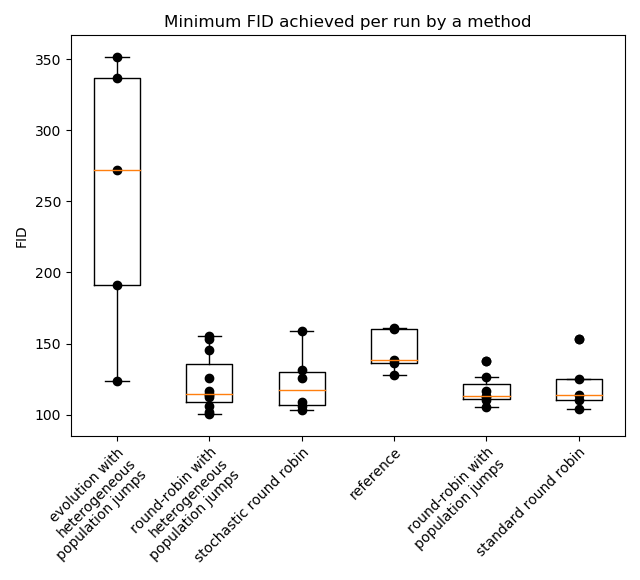}
\caption{Boxplot of best FIDs achieved per run}
\label{fig:fid_best_box}
\end{center}
\end{figure}

\begin{figure}[!htb]
\begin{center}
\includegraphics[width=3.0in]{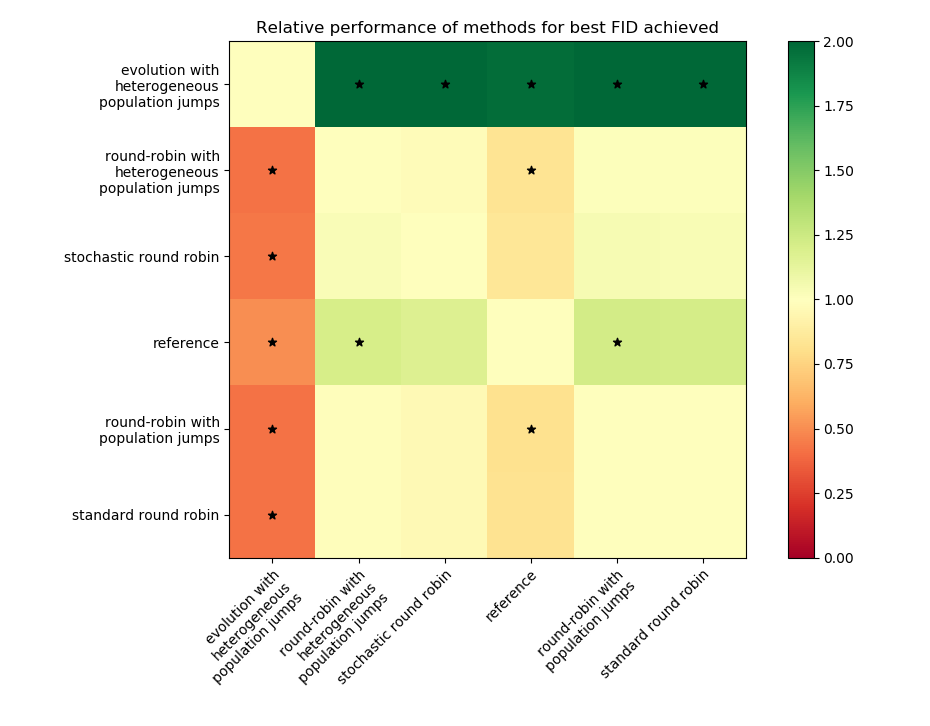}
\caption{Comparison of best FIDs per run. Color indicates the relative performance of the method on the line compared to the method on the column (median ratio). Green is better, asterisk indicates statistically significant difference.}
\label{fig:fid_best_comp}
\end{center}
\end{figure}

\begin{figure}[!htb]
\begin{center}
\includegraphics[width=3.0in]{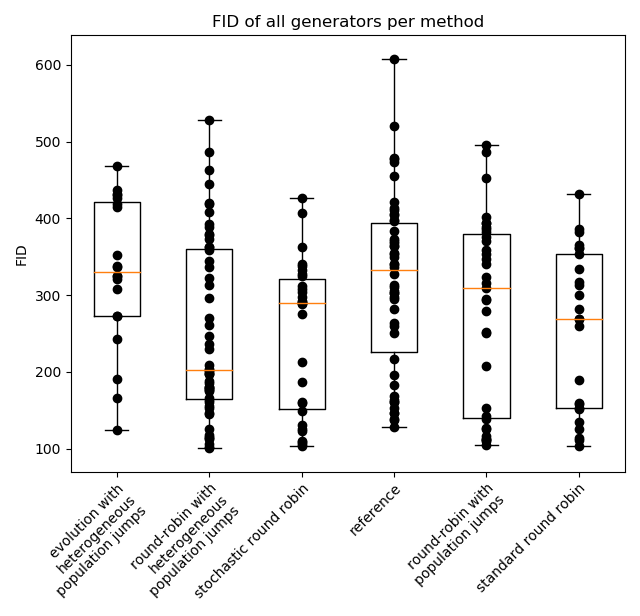}
\caption{Boxplot of FIDS achieved by all generators per method.}
\label{fig:fid_all_box}
\end{center}
\end{figure}

\begin{figure}[!htb]
\begin{center}
\includegraphics[width=3.0in]{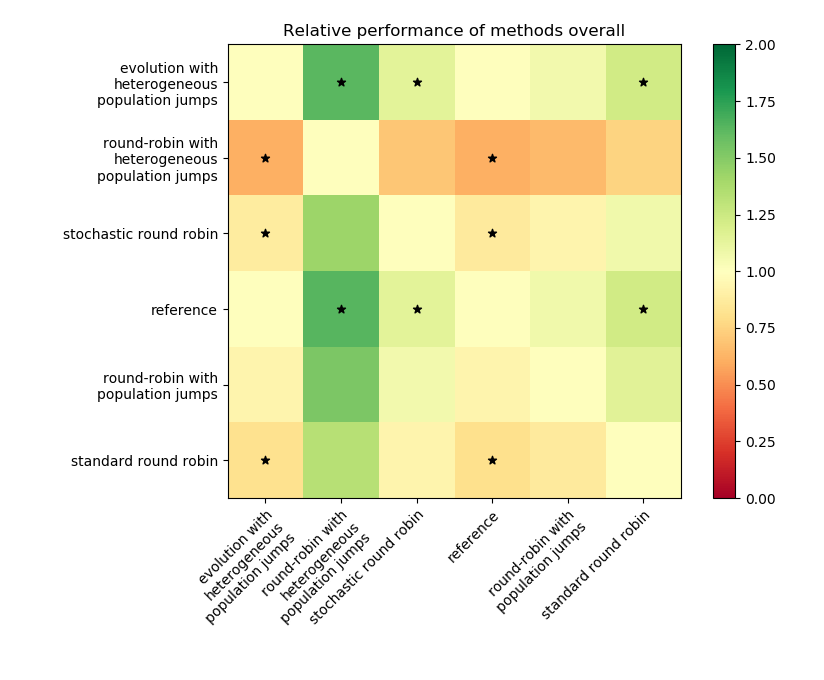}
\caption{Comparison of all FIDs achieved. Color indicates the relative performance of the method on the line compared to the method on the column (median ratio). Green is better, asterisk indicates statistically significant difference.}
\label{fig:fid_all_comp}
\end{center}
\end{figure}

The summary of the best FID statistics achieved can be seen in the table \ref{tab:fid_stats} and in Fig.\ref{fig:fid_best_box}. Comparison of method performances can be found in Fig.\ref{fig:fid_best_comp}. Similarly, the statics for the FIDs of all the generators in in the methods' runs can be found in Figs. \ref{fig:fid_all_box} and \ref{fig:fid_all_comp}.

\begin{table}[!hbt]
\center{
\begin{tabular}{| m{7em} | m{1cm} | m{1cm} | m{1cm} |} \hline
propagation structure & median FID & mean FID & FID std \\ \hline\hline
evolution with
heterogeneous
population jumps & 272.35 & 255.04 & 86.77 \\ \hline
round-robin with
heterogeneous
population jumps & 114.62 & 122.45 & 19.12 \\ \hline
stochastic round robin & 117.36 & 122.43 & 19.22\\ \hline
reference & 138.28 & 144.72 & 13.43\\ \hline
round-robin with
population jumps & 112.89 & 117.33 & 10.39 \\ \hline
standard round robin & 113.52 & 121.23 & 17.49 \\ \hline
\end{tabular}
\vskip 0.25cm
\caption{Best FID statistics achieved per run for each method}
\label{tab:fid_stats}
}
\end{table}

The data here suggests several conclusions.

\subsection{Round-Robin with heterogeneous populations jumps outperforms all other matching methods.}

Just as expected, it is clear from Figs. \ref{fig:fid_best_comp} and \ref{fig:fid_best_box} that Round-Robin with heterogeneous population jumps significantly improves over baseline even despite reduced computational training budget. Similarly, as Figs. \ref{fig:fid_all_box} and \ref{fig:fid_all_comp} show, for the median FID achieved per run, it outperforms any other training method.
This agrees with our hypothesis that both jumps between naive populations and heterogeneity of the discriminators between naive populations seem to play a role in allowing all training FIDs to keep decreasing. 

\begin{figure}[!htb]
\begin{center}
\includegraphics[width=3.0in]{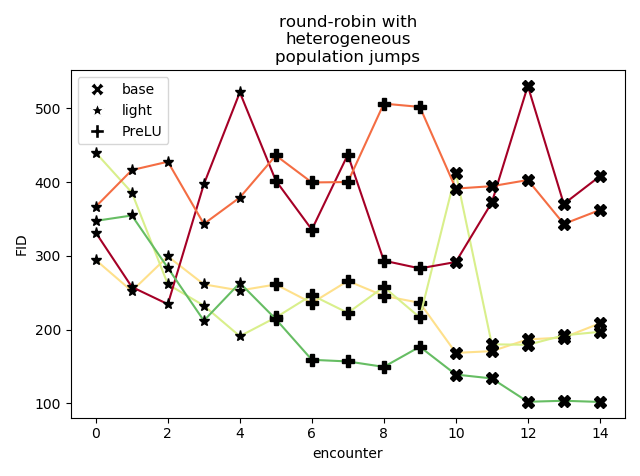}
\caption{A typical run for a heterogeneous population jumping round-robin method.}
\label{fig:rr_het_pop}
\end{center}
\end{figure}

Fig. \ref{fig:rr_het_pop}, representing a typical run, confirms our initial prediction that such a training regimen preserves potentially adaptive heterogeneity of the pathogen original population and thanks to the switching between the environments forces a longer adaptive march, preventing the generators from remaining stuck in local minimums \footnote{Most likely, new discriminators would anchor their discriminatory capacity in the features of the generated images that have been overlooked by the previous hosts populations discriminators.}.

\subsection{Other versions of Round-Robin training perform surprisingly well.}

While Round-Robin with heterogeneous populations jumps achieves lowest average FIDs, the comparison of all the Round-Robin training methods with regard to the best FID achieved per run is surprisingly close. They are all robustly outperform the reference method, despite some of them having a computational budget that is only half that of the reference.

\begin{figure}[!htb]
\begin{center}
\includegraphics[width=3.0in]{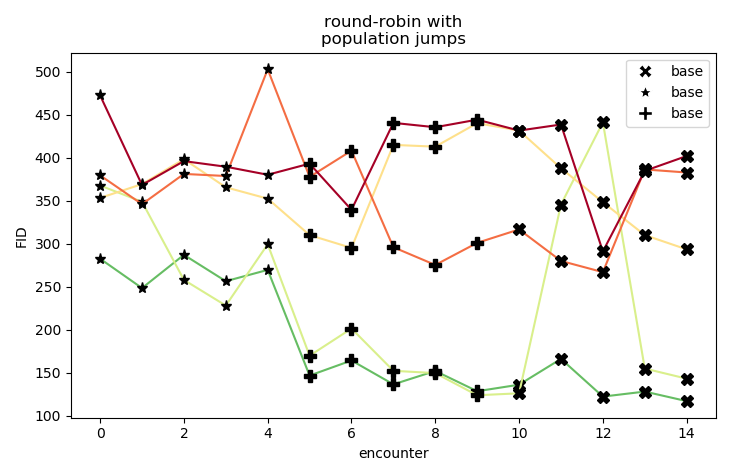}
\caption{A typical run for a homogeneous population jumping Round-Robin method.}
\label{fig:rr_hom_pop}
\end{center}
\end{figure}

\begin{figure}[!htb]
\begin{center}
\includegraphics[width=3.0in]{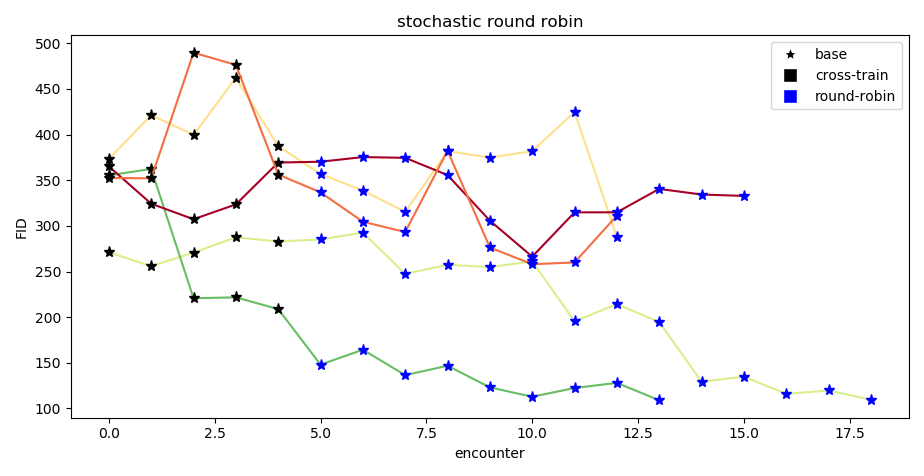}
\caption{A typical run for a stochastic, single population Round-Robin.}
\label{fig:rr_stoch}
\end{center}
\end{figure}

It seems that the diversity of the discriminators encountered by the generators, even in Round-Robin implementation without population jumps, is sufficient to keep adaptive swipes going, even if it is likely that with smaller change in the environment without heterogeneous populations, the neighboring optima search is less efficient on average. This is illustrated by figures Fig. \ref{fig:rr_hom_pop} and \ref{fig:rr_stoch}.

\subsection{Evolutionary phase seems to negate the advantages of Round-Robin based training methods and population jumps.}

\begin{figure}[!htb]
\begin{center}
\includegraphics[width=3.0in]{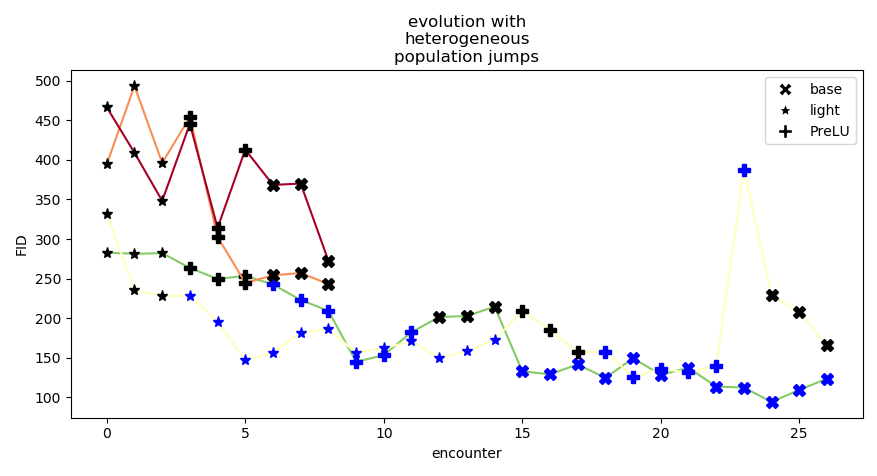}
\caption{A typical run for the evolutionary method.}
\label{fig:evo}
\end{center}
\end{figure}

Finally, an unexpected result of our simulation is that with the structure of selection process and fitness functions we've chosen here, the evolutionary phase does not seem to improve the outcome. Given the FID evolution over a typical run, as illustrated in Fig. \ref{fig:evo}, we can hypothesize that this is due to the fact that the best performing discriminator is recurrently picked and trained, leading to the fitness collapse of most generators and the same issues as present in the reference single generator/single discriminator GAN training method.

While unexpected, this is hardly surprising. First, the design of evolutionary architectures that do not require laboriously defined fitness functions is a non-trivial task, requiring to make sure that the pressure imparted by the selection is consistent with the testing environment. Second, as we mentioned in the introduction, the training of GANs is already a proxy for evolutionary process and a secondary one is likely interfering with it rather than enhancing it. Exploring such an interaction and designing an evolutionary architecture that would take advantage of it is an exciting potential direction of future research, but is unfortunately outside the scope of this study.

\section{Discussion}

In this paper, we present a novel perspective on GANs training - that of host-pathogen co-evolutionary theory. We show that while dissimilar to the well-known genetic algorithm (eg. \cite{whitley1994genetic}, the more recent pathogen-host co-evolution theory is appropriate in the context of GAN training. We are able to exploit more recent advances in evolutionary theory and using it, suggest more computationally efficient methods of training GANs and experimentally show that they can relatively rapidly lead to the emergence of performant adversarial generative networks.

Interestingly, while the reasons for the performance of our methods come from the evolutionary perspective, the training architecture suggested here can be further improved upon by using other insights from host-pathogen co-evolution that have theoretical pendants in the GAN training theory. For instance, accounting for the larger size of populations and longer time needed to reach herd immunity would lead to a slower speed of discriminator training compared to that of a generator - consistent with the Two Time-Scale Update Rule suggested in \cite{heusel2017gans}.

In a similar vein, the previously suggested ways of improving the GAN training process also explain why we rarely see a stalling in the host-pathogen co-evolution due to the over-performance of the host immune system similar to the ones plaguing GANs. 
Unlike biological organisms, GANs are susceptible to the measure collapse \cite{mahloujifar2019curse, arora2017generalization}, when the manifold of samples that the generator is capable of generating is disjointed from the manifold of real samples and a sufficiently complex discriminator can learn to distinguish the two. Remedies that has been suggested for that - imprecise matching and Gaussian noise - are naturally present in biological context. Antibodies present a broad spectrum of sensitivities to protein 3D configurations, recognizing not only the exact shapes but also similar ones. Similarly, due to the Brownian motion, epitope surfaces of the pathogens do not have a strictly defined boundary or shape - instead they present a spatial distribution of positions and properties (such as electronegativity or polarity). A competent immune system will need to learn to distinguish similar probability fields generated by the epitopes of the pathogen or those of the host organism. Including imprecise matching is outside the scope of this paper, but is a promising avenue of future research.

Another interesting venue of future investigation would be to see if insights from GANs training processes could be of any use to improve our understanding of the biological and evolutionary processes involved in pathogen-host co-evolution. While an exciting perspective, this venue of research requires accounting for several crucial differences needed to generalize the results of experiments on GANs back to host-pathogen pairs. 

First, in the living organisms, the genome and the phenotype are two distinct entities and the relationship between them is not that direct. A considerable amount of theoretical work indicates that the evolvability (how few mutations are needed to provide drastic adaptive modifications of the phenotype) is itself under evolutionary selection. As such we expect that the biological organisms evolve more rapidly, using a "bag of tricks" learned through their evolutionary history and encoded in the mapping between the genome and the phenotype. The relation between neural network weights by gradient back-propagation in GANs is unlikely to be consistent with such mapping.

Second, the actual evolution does not occur by gradient descent, but rather by random mutations, some of which may reveal themselves beneficial now, but a lot of which would be neutral - potentially beneficial in the future, upon the environment change. In case of viruses, there are also massive constraints to how the genome can evolve, that are due to the geometry of the viral capside side and usage of the cell receptors for the entry into the host cells. Such limitations are, once again, not present in the standard GANs. 

Finding how such limitations can be encoded into the GAN architecture could open a venue for exploring interesting analogies. 

For instance, for a number of rapidly mutating RNA-based viruses, the high mutability of virus leads to a non-convergence to a single optimal viral sequence, but rather an ergodic oscillation around an optimum in the genetic and phenotypic spaces. In GANs world that has an immediate parallel of non-convergence of GAN. 

As another example, similarly to highly efficient immune system stopping the virus' evolution in its tracks, a high-quality discriminator will lead to vanishing gradients. Some viruses exploit specific features of the immune system, such as HIV exploiting the T-cells and their role in human immune system - similarly how some generators can exploit non-robust features specific to some discriminator architectures.

Such similarity in behaviours suggests some core common principles of design processes between GANs and pathogen-host dynamics. As such, the parallels between the GANs training theory and pathogen-host co-evolution deserves a more in-depth investigation in the future.

\section{Acknowledgements}

This work was supported by the Swiss National Science Foundation grant N 200021\_182542/1.

\footnotesize
\bibliographystyle{acm}
\bibliography{biblio}

\end{document}